\documentclass[10pt,twocolumn,letterpaper]{article}

\usepackage{iccv}              

\definecolor{iccvblue}{rgb}{0.21,0.49,0.74}
\usepackage[pagebackref,breaklinks,colorlinks,allcolors=iccvblue]{hyperref}
\usepackage{amsmath,amsfonts}
\usepackage{algorithmic}
\usepackage{algorithm}
\usepackage{array}
\usepackage{textcomp}
\usepackage{stfloats}
\usepackage{url}
\usepackage{verbatim}
\usepackage{graphicx}
\usepackage{color}
\usepackage{appendix}
\usepackage{multirow}
\usepackage{multicol}
\usepackage{multirow}
\usepackage{algorithm}
\usepackage{algorithmic}
\usepackage{booktabs} 
\usepackage{multirow}
\usepackage{placeins}
\usepackage{amsfonts}
\usepackage{makecell}
\def\paperID{2548} 
\def\confName{ICCV}
\def\confYear{2025}

\title{ONER: Online Experience Replay for Incremental Anomaly Detection}

\author{
Yizhou Jin, Jiahui Zhu, Guodong Wang, Shiwei Li, Jinjin Zhang, Xinyue Liu, Qingjie Liu$\thanks{Corresponding Author.}$, Yunhong Wang\\
\textsuperscript{\rm 1}State Key Laboratory of Virtual Reality Technology and Systems, Beihang University, Beijing, China\\
\textsuperscript{\rm 2}Hangzhou Innovation Institute, Beihang\\
}
\begin{document}
\maketitle
\begin{abstract}
Incremental anomaly detection aims to sequentially identify defects in industrial product lines but suffers from catastrophic forgetting, primarily due to knowledge overwriting during parameter updates and feature conflicts between tasks. In this work, We propose \textbf{ONER} (\textbf{ON}line \textbf{E}xperience \textbf{R}eplay), an end-to-end framework that addresses these issues by synergistically integrating two types of experience: (1) decomposed prompts, which dynamically generate image-conditioned prompts from reusable modules to retain prior knowledge thus prevent knowledge overwriting, and (2) semantic prototypes, which enforce separability in latent feature spaces at pixel and image levels to mitigate cross-task feature conflicts. Extensive experiments demonstrate the superiority of ONER, achieving state-of-the-art performance with +4.4\% Pixel AUROC and +28.3\% Pixel AUPR improvements on the MVTec AD dataset over prior methods. Remarkably, ONER achieves this with only 0.019M parameters and 5 training epochs per task, confirming its efficiency and stability for real-world industrial deployment.
\end{abstract}
\section{Introduction}\label{sec:introduction}
Anomaly detection (AD) in computer vision aims to identify rare patterns or outliers that significantly deviate from the normal data distribution. By learning exclusively from normal data, AD plays a critical role in industrial applications, where anomalies are rare, diverse, and costly to annotate.

However, most existing AD methods~\cite{DBLP:conf/iccv/ZavrtanikKS21, DBLP:conf/kdd/AudibertMGMZ20,Bergmann_Löwe_Fauser_Sattlegger_Steger_2019,DBLP:conf/aaai/YanZXHH21, DBLP:journals/corr/abs-2110-03396,liang2023omni,DBLP:conf/cvpr/WyattLSW22, DBLP:conf/aaai/HeZCCLCWW024,Yao_Liu_Wang_Yin_Yan_Hong_Zuo_2024,DBLP:conf/cvpr/SalehiSBRR21,DBLP:conf/bmvc/WangHD021,DBLP:conf/accv/YiY20, DBLP:conf/cvpr/LiSYP21, DBLP:journals/eaai/YangWF23, DBLP:conf/cvpr/ReissCBH21,rudolph2021same, DBLP:conf/wacv/RudolphWRW22, DBLP:journals/corr/abs-2206-01992} focus on single-product scenarios, requiring separate models per product and specific product identity ahead of testing. Although effective in a controlled setting, this product-specific design remains with two critical industrial limitations: 1) \textbf{Incapability to handle multiple types of product} on mixed production lines, and 2) \textbf{Inflexibility to dynamic schedule changes}~\cite{towill1997analysis}. Furthermore, training separate models per product introduces excessive training, deployment, and storage overheads, particularly in large-scale operations where resource efficiency is paramount. These limitations severely restrict deployment in dynamic industrial settings where product lines evolve rapidly. Thus, there is a pressing need for a model capable of rapid adaptation to new tasks, such as detecting new products, with high detection accuracy while maintaining historical task performance.

\begin{figure}[tp]
    \centering
    \includegraphics[width=0.92\linewidth]{
    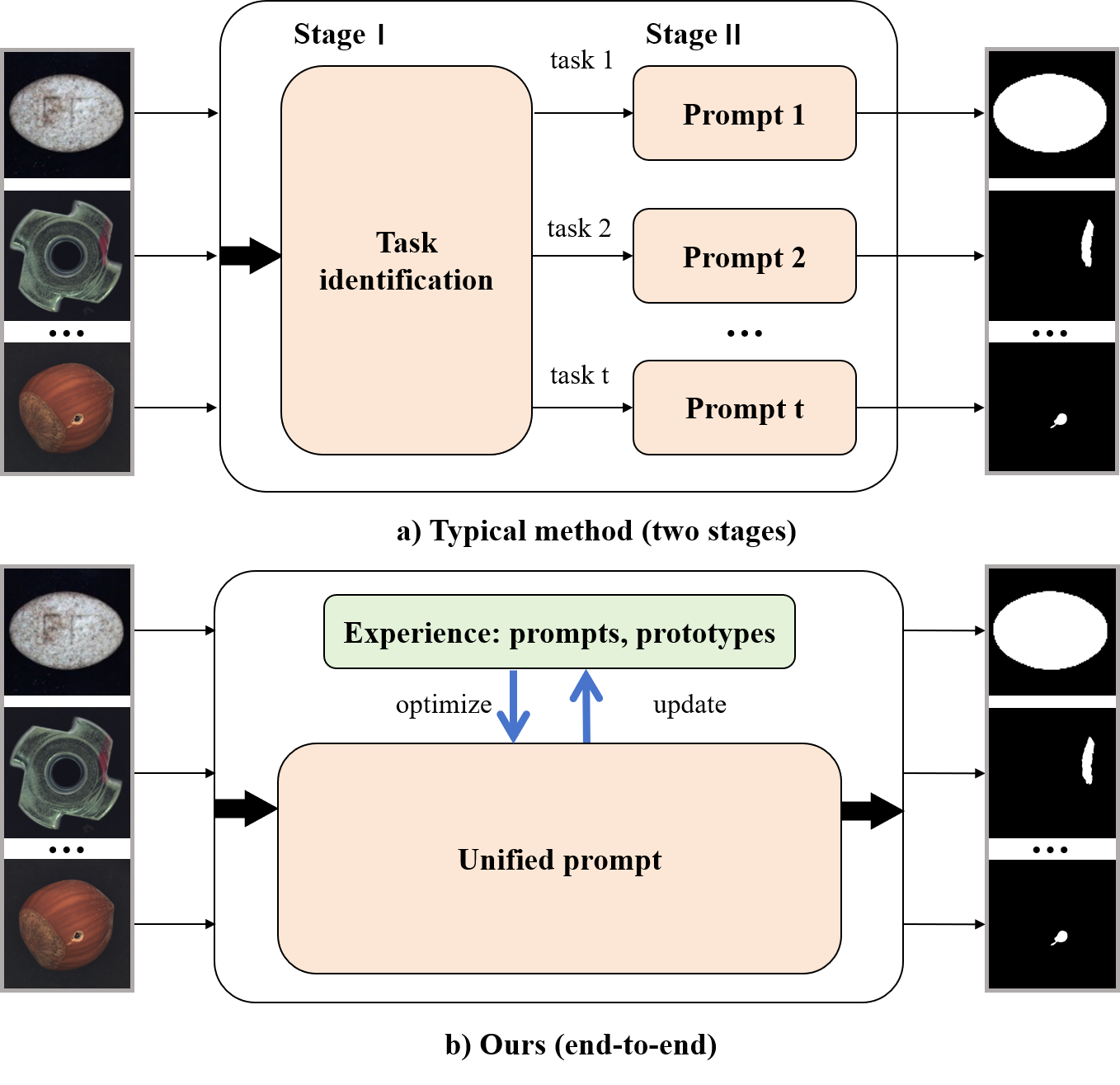}
    \caption{Comparison between a typical incremental AD method and ours: (a) Typical methods train task-specific prompts and require task identification during inference. (b) We employ an end-to-end prompt for both training and inference, bypassing the need for task identification. This avoids staged processing, minimizes cumulative errors, and enables seamless adaptation to new tasks while preserving prior knowledge by leveraging past experience effectively.} 
    \label{fig:framework-IUAD}
\end{figure}

Training-free AD methods~\cite{Roth_2022_CVPR,kim2023fapm,kim2021semi,cohen2020sub,Li_Ivanova_London,jeong2023winclip,cao2023segment,li2023clip,chen2024clip,Zhou_Pang_Tian_He_Chen_2024,qu2024vcp} bypass task-specific parameter optimization, inherently avoiding catastrophic forgetting and reducing training overheads. Although these approaches achieve reasonable performance on average, they are unable to detect subtle anomalies due to the absence of task-specific knowledge, rendering them inadequate for real-world incremental industrial applications.

Recent research on AD has increasingly adopted incremental learning to meet the growing demand for adaptability in dynamic manufacturing environments.

CAD~\cite{li2022towards} first proposes an incremental AD framework that uses a Gaussian distribution estimator. It proposes leveraging the feature distribution of normal training samples from previous tasks to effectively alleviate catastrophic forgetting. However, incremental optimization in CAD is conducted on the classifier head, which enables anomaly classification but does not support pixel-level anomaly segmentation.

Inspired by prompt-based approaches~\cite{Wang_Zhang_Lee_Zhang_Sun_Ren_Su_Perot_Dy_Pfister_2022,wang2022dualprompt} that utilize the generalization capabilities of pretrained vision transformers~(ViTs)~\cite{dosovitskiy2020image}, UCAD~\cite{liu2024unsupervised} can adapt to incremental tasks by introducing a continual prompting module. Despite favorable results achieved, it suffers from notable limitations, as shown in Fig.~\ref{fig:framework-IUAD}~(a). UCAD adopts a staged learning pipeline, training separate prompts for each task, and relying on a task-specific inference process. This process requires identifying the task associated with each input image before applying the corresponding prompt for AD. However, this dependence on task identification is prone to cumulative errors, as incorrect prompts can lead to complete detection failure. While the follow-up work IUF~\cite{tang2024incremental} avoids cumulative errors  by introducing an end-to-end learning framework, it suffers from high training costs, \textit{e.g.}, 200 epochs per task, making it less feasible for real-world scenarios. 

To address these challenges, we propose an end-to-end \textbf{ON}line \textbf{E}xperience \textbf{R}eplay (ONER) approach, as illustrated in Fig.~\ref{fig:framework-IUAD} (b). ONER mitigates catastrophic forgetting, a phenomenon primarily caused by knowledge overwriting during parameter updates \cite{xu2024mitigate} and feature conflicts between tasks~\cite{tang2024incremental}. Simultaneously, it adapts to new tasks with minimal training overheads through two key components: decomposed prompts and semantic prototypes, which jointly optimize model parameters and feature representations. Decomposed prompts reuse frozen parameters from prior tasks while introducing learnable parameters for new tasks, thereby preventing knowledge overwriting. By generating end-to-end image-conditioned prompts, ONER avoids cumulative errors common in staged approaches~\cite{liu2024unsupervised}, ensuring stable anomaly detection. Semantic prototypes store pixel- and image-level features from historical tasks, enforcing regularization by maximizing feature distances between past and current tasks. This mitigates feature conflicts and enhances discriminability both within and across tasks.

In summary, our main contributions are listed as follows:
 \begin{itemize}
 \item We propose ONER, an end-to-end online experience replay framework that efficiently mitigates catastrophic forgetting while adapting to new tasks with minimal training oveerheads (0.019M learnable parameters, 5 epochs per task).
\item To address catastrophic forgetting caused by knowledge overwriting in parameter updates and cross-task feature conflicts, ONER incorporates two types of experience: decomposed prompts and semantic prototypes, jointly addressing model parameter updates and feature optimization. 
\item  Extensive experiments demonstrate the superiority of ONER, achieving state-of-the-art performance with +4.4\% Pixel AUROC and +28.3\% Pixel AUPR improvements on the MVTec AD dataset over prior methods.

\end{itemize}
\section{Related Work}
\subsection{Anomaly Detection}
Existing methods for anomaly detection can generally be divided into two categories: reconstruction-based and feature-embedding \cite{LiuXWLWZJ24}.
Reconstruction-based methods utilize neural networks for data encoding and decoding, implicitly learning the normal sample distribution. During training, the network is trained on normal samples, which facilitates the identification of normal characteristics but may diminish its efficacy in reconstructing anomalies. Anomalies are identified during the testing phase through a discrepancy analysis between the original input and the reconstructed output. Various network architectures, including autoencoders \cite{DBLP:conf/iccv/ZavrtanikKS21, DBLP:conf/kdd/AudibertMGMZ20,Bergmann_Löwe_Fauser_Sattlegger_Steger_2019}, GANs \cite{DBLP:conf/aaai/YanZXHH21, DBLP:journals/corr/abs-2110-03396,liang2023omni}, and diffusion models \cite{DBLP:conf/cvpr/WyattLSW22, DBLP:conf/aaai/HeZCCLCWW024,Yao_Liu_Wang_Yin_Yan_Hong_Zuo_2024}, are applied to the task of anomaly detection.
Feature-embedding methods meticulously extract distinctive features, thereby amplifying detection precision by discarding extraneous attributes. One-class classification techniques \cite{DBLP:conf/accv/YiY20, DBLP:conf/cvpr/LiSYP21, DBLP:journals/eaai/YangWF23, DBLP:conf/cvpr/ReissCBH21} employ hyperspheres to differentiate features. Distribution mapping strategies \cite{rudolph2021same, DBLP:conf/wacv/RudolphWRW22,  DBLP:journals/corr/abs-2206-01992} project features onto Gaussian distributions, facilitating the evaluation of anomalies. 

\subsection{Training-free Anomaly Detection}
A common strategy for training-free AD involves extracting normal product features via a robust pre-trained network and storing them in a memory bank. During inference, anomalies are detected by comparing the features of the test image with the memory bank and calculating the spatial distances to the normal features. However, these methods~\cite{Roth_2022_CVPR,kim2023fapm,kim2021semi,cohen2020sub} remain task-specific, as they require separate memory banks for each category of products and explicit selection during testing. To overcome this limitation, some approaches~\cite{Li_Ivanova_London,jeong2023winclip,cao2023segment,li2023clip} integrate text encoders (e.g. CLIP~\cite{radford2021learning}), performing anomaly detection by aligning visual features with predefined textual prompts (e.g. "normal" vs. "abnormal"). Although this reduces product identity dependency, it struggles with granular anomalies due to semantic gaps between generic text prompts and domain-specific defects. Recent efforts~\cite{chen2024clip,Zhou_Pang_Tian_He_Chen_2024,qu2024vcp} further fine-tune models with external multimodal datasets to enhance fine-grained visual-semantic alignment, but this compromises the training-free principle by reintroducing parameter updates. 
\begin{figure*}[!ht]
    \centering
    \includegraphics[width=0.95\linewidth]{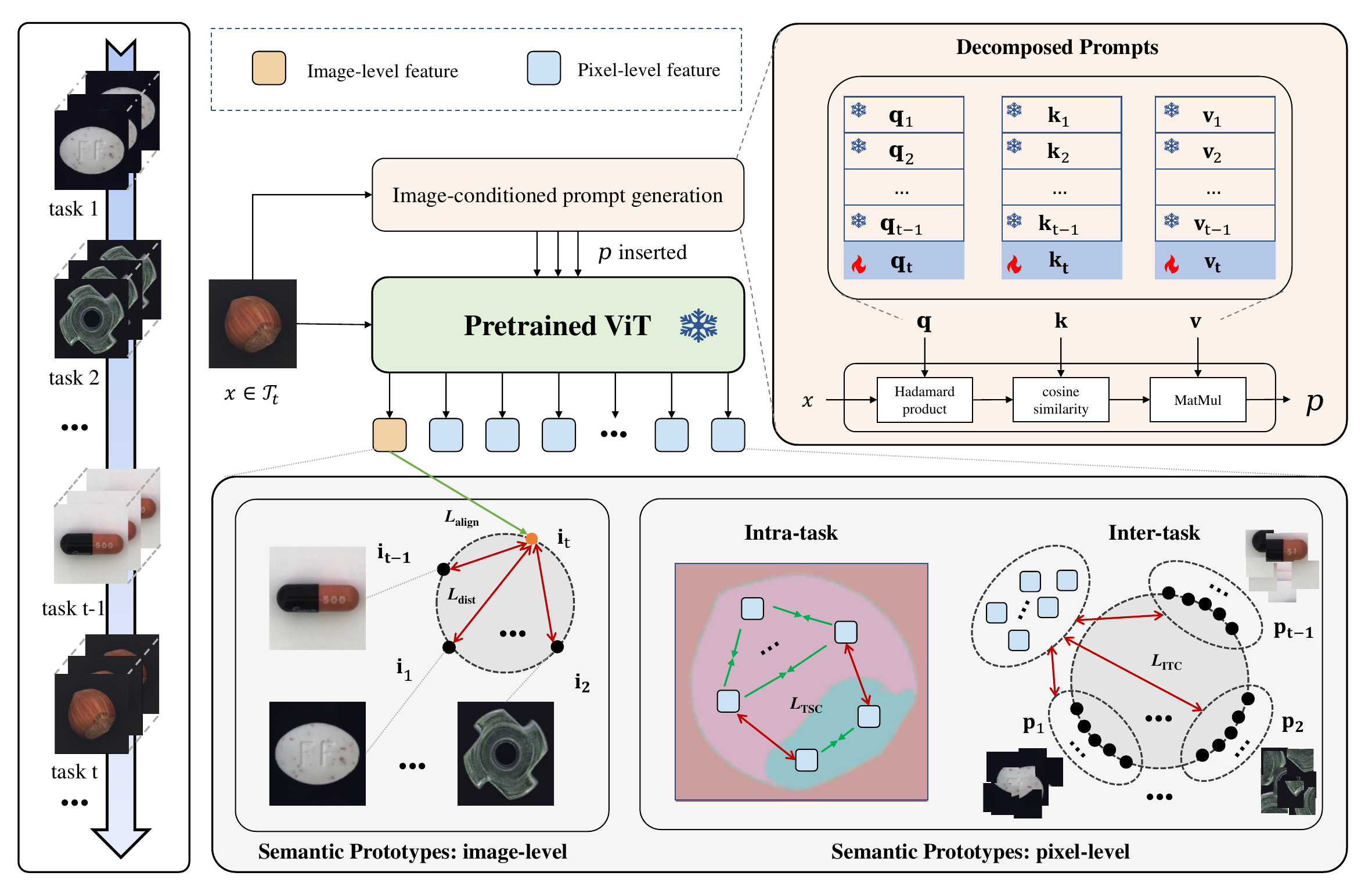}
    \caption{This diagram illustrates the structure of ONER. Specifically, ONER incorporates two types of experience: decomposed prompts and semantic prototypes. Decomposed prompts, formed from learnable components, reuse previous knowledge to help the model learn new tasks. Semantic prototypes provide regularization at both pixel and image levels, preventing forgetting across tasks. The diagram shows the training process for step $t$, where both the decomposed prompts and semantic prototypes are updated after training.} 
    \label{fig:framework}
\end{figure*}
\subsection{Incremental Anomaly Detection}
Incremental learning-based anomaly detection algorithms can adapt to novel tasks without retraining, and thus enable the quick detection of emerging anomalies.
Since obtaining labeled dataset for unforeseen defects poses a significant challenge within the context of industrial manufacturing, unsupervised anomaly detection with incremental training is crucial.
LeMO \cite{DBLP:journals/corr/abs-2305-15652} follows traditional unsupervised anomaly detection but does not address inter-class incremental anomaly detection in rapidly expanding industrial product categories.
To address this issue, CAD \cite{DBLP:conf/mm/LiZWXGLWZ22} introduces an unsupervised feature embedding task for continuous learning. 
This approach recommends the encapsulation of multiple object features within a unified storage framework, with defect identification facilitated by a binary classification mechanism.
However, this method is limited to classification and cannot be used for segmentation. IUF \cite{tang2024incremental} combines a unified model for multiple objects with object incremental learning, enabling pixel-accurate defect inspection across different objects without feature storage in memory. However, this method lacks an explicit representation of the semantic space, hindering the measurement of the network's ability in object incremental learning. UCAD \cite{DBLP:conf/aaai/0004WNCGLWWZ24} proposes continuous learning by establishing a key prompt knowledge memory space. However, UCAD requires separate training and storage for each category, and during testing, it confirms the category of the input image before testing, which is less efficient.

\section{Methods}
\subsection{Problem Formulation}
In our incremental anomaly detection framework, the training and testing datasets are divided into $N$ distinct subsets, each corresponding to a distinct product category. The training set $\mathcal{T} = \{\mathcal{T}_1, \mathcal{T}_2, \ldots, \mathcal{T}_N\}$ contains normal samples only, whereas the test set $\mathcal{T}^{\prime} = \{\mathcal{T}^{\prime}_1, \mathcal{T}^{\prime}_2, \ldots, \mathcal{T}^{\prime}_N\}$ includes mixed normal and anomalous samples.

The model undergoes $N$ sequential training steps, incrementally learning from each subset $\mathcal{T}_t$, $t \in [1, N]$. Unlike frameworks that employ sample replay, our model operates in a strict rehearsal-free manner, prohibited from revisiting training data from previous steps~(i.e. steps 1 to $t -1$). After full incremental training, the final model is evaluated on the entire test set $\mathcal{T}^{\prime}$, performing anomaly detection without product identity priors.

\subsection{Overview}
In incremental learning, effectively leveraging experience from previous tasks is critical for mitigating catastrophic forgetting~\cite{rolnick2019experience}. This phenomenon primarily stems from two factors: knowledge overwriting~\cite{xu2024mitigate} during parameter updates and feature conflicts
~\cite{tang2024incremental} between tasks. To address these challenges, our framework introduces two core experience components~(shown in Fig. \ref{fig:framework}): \textbf{decomposed prompts} $\mathcal{D}$ and \textbf{semantic prototypes} ($\mathcal{I},\mathcal{P}$), which jointly optimize model parameters and feature representations through a corresponding memory bank.

Decomposed prompts $\mathcal{D}=\{\mathbf{d}_{1}, \mathbf{d}_{2}, \ldots, \mathbf{d}_\text{N}\}$ enable parameter-efficient knowledge transfer by reusing frozen parameters from prior tasks while expanding learnable components for new tasks. Inspired by \cite{smith2023coda}, each prompt $\mathbf{d}_\text{i}=\{\mathbf{q}_\text{i},\mathbf{k}_\text{i},\mathbf{v}_\text{i}\}$ consists of three learnable matrices: attention queries $\mathbf{q} \in \mathbb{R}^{M \times d}$ and keys $\mathbf{k}\in \mathbb{R}^{M \times d}$ for task-specific attention conditioning, while prompt values $\mathbf{v}\in \mathbb{R}^{M\times L_p \times d}$ storing prompt embeddings. Here $d$ is the embedding dimension, $L_p$ the prompt length, and $M=M_c\times N$ the number of prompt components~($M_c$: new components per task). When task $t$ arrives, $M_c$ task-specific components $\mathbf{d}_{\text{t}^*}=\{\mathbf{q}_{\text{t}^*},\mathbf{k}_{\text{t}^*},\mathbf{v}_{\text{t}^*}\}$ are added to the prompt bank $\mathcal{D}_{t^*}\gets\mathcal{D}_{t-1}\cup \{\mathbf{d}_{\text{t}^*}\}$, with prior components $\mathcal{D}_{t-1}$ frozen. Only $\mathbf{d}_{\text{t}^*}$ is optimized, preventing historical knowledge overwriting.

Semantic prototypes preserve task-invariant representations through two complementary banks. Image-level prototypes $\mathcal{I}=\{\mathbf{i}_1, \mathbf{i}_2,\ldots, \mathbf{i}_\text{N}\}$ encode global normal patterns via incremental refinement. Pixel-level prototypes $\mathcal{P}=\{\mathbf{p}_1, \mathbf{p}_2,\ldots, \mathbf{p}_\text{N}\}$ capture local discriminative features.

As outlined in Algo.~\ref{alg:train}, training proceeds in $N$ sequential steps. 
When task $t$ arrives, the \textbf{ Image-level Prototypes Refinement (IPR)} module process $\mathcal{I}_{t-1}$ and $\mathcal{T}_t$ to generate $\mathbf{i}_\text{t}$, updating the bank as $\mathcal{I}_{t}\gets\mathcal{I}_{t-1} \cup \{\mathbf{i}_\text{t}\}$. Concurrently, the prompt bank expands to $\mathcal{D}_{t^*}$, ensuring parameter stability.

During training, a prompt $p$ is synthesized for each image $x\in\mathcal{T}_t$ via attention-weighted aggregation:
\begin{equation}
p = \alpha \mathbf{v}, \quad \alpha = \gamma\left(q(x), \mathbf{k}\right), \quad q(x) = f_\theta(x) \odot \mathbf{q},
\end{equation}
where $\gamma$ computes cosine similarity and $f_\theta$ is a pre-trained ViT.

The prompt $p$ is injected into the layers of $f_\theta$, forming an adapted model $f_\phi(x)=f_{\theta, \mathcal{D}}(x)$ to steer feature extraction. The extracted features $\mathbf{k}$ are divided into image-level $\mathbf{k}_\text{I}\in\mathbb{R}^{1 \times d}$ and pixel-level $\mathbf{k}_\text{P}\in \mathbb{R}^{N_p \times d}$ components, where $N_p$ is the number of patches. These features, along with $\mathcal{I}_{t}$ and retrieved $\mathcal{P}_{t-1}$, are then processed by the \textbf{ Experience Replay-based Optimization (ERO)} module to optimize $\mathcal{D}_{t^*}$, yielding updated prompts $\mathcal{D}_{t}$.
  
Upon completing training for task $t$, the \textbf{Incremental Selection of  Pixel-level Prototypes~(ISPP)} module selects representative pixel-level prototypes $\mathbf{p}_\text{t}$ from $ \mathcal{T}_t$, merging them into the global bank as $\mathcal{P}_{t} \gets \mathcal{P}_{t-1}  \cup \{\mathbf{p}_\text{t}\}$.

This dual mechanism, as decomposed prompts for model optimization and semantic prototypes for feature representation, ensures progressive knowledge integration while systematically mitigating forgetting.

\begin{algorithm}[!t]
\caption{Training procedure}
\label{alg:train}
\renewcommand{\algorithmicrequire}{\textbf{Input:}}
\renewcommand{\algorithmicensure}{\textbf{Output:}}
\definecolor{codeblue}{rgb}{0.25,0.5,0.5}
\textcolor{codeblue}{\# The overall training process is divided into N steps, for each step $t\in [1, N]$:}
\begin{algorithmic}[1]
\REQUIRE{
    Retrieved experience: $\mathcal{D}_{t-1}$,
    $\mathcal{I}_{t-1}$,
    $\mathcal{P}_{t-1}$, training datasets $\mathcal{T}_t \in \{\mathcal{T}_1, \mathcal{T}_2, \ldots, \mathcal{T}_N\}$, pretrained ViT $f_\theta$
}\\
\ENSURE{
   updated experience: $\mathcal{D}_{t}$,
    $\mathcal{I}_{t}$,
    $\mathcal{P}_{t}$
}\\
    \STATE \textcolor{codeblue}{\# Pre-stage:}
    \STATE  \textbf{expand} $\mathcal{D}_{t^*}\gets \mathcal{D}_{t-1} \cup  \{\mathbf{d}_{\text{t}^*}\}$
    \STATE $\mathbf{i}_\text{t} \gets \textbf{IPR}( \mathcal{I}_{t-1}, \mathcal{T}_t)$ \textcolor{codeblue}{\# refers to section 3.3}
    \STATE \textbf{update} $\mathcal{I}_{t} \gets \mathcal{I}_{t-1} \cup \{\mathbf{i}_\text{t}\}$
    \STATE \textcolor{codeblue}{\# Training stage:}
    \WHILE{training not converged}
     \FOR{batch $x \in \mathcal{T}_t$}
    \STATE $\mathbf{k}_\text{I},\mathbf{k}_\text{P} \gets f_{\theta, \mathcal{D}_{t^*}}(x)$
    \STATE $\mathcal{L}_\text{align},\mathcal{L}_\text{TSC},\mathcal{L}_\text{ITC} \gets \textbf{ERO}(\mathcal{I}_{t},\mathbf{k}_\text{I},\mathcal{P}_{t-1},\mathbf{k}_\text{P})$ \textcolor{codeblue}{\# refers to section 3.4}
    \STATE $\mathcal{L}_\text{total} =\mathcal{L}_\text{align} + \mathcal{L}_\text{TSC} + \mathcal{L}_\text{ITC}$
    \STATE $ \mathcal{D}_{t^*} \gets \textbf{BP}(\mathcal{L}_\text{total})$
    \ENDFOR
    \ENDWHILE
   
    \STATE \textcolor{codeblue}{\# Post-stage:}
    \STATE \textbf{fix} $\mathcal{D}_{t} \gets \mathcal{D}_{t^*}$
    \STATE $\mathbf{p}_\text{t} \gets \textbf{ISPP}(\mathcal{T}_t,\mathcal{D}_{t},\mathcal{P}_{t-1})$ \textcolor{codeblue}{\# refers to section 3.5}
    \STATE \textbf{update} $\mathcal{P}_{t} \gets \mathcal{P}_{t-1} \cup \{\mathbf{p}_\text{t}\}$
\STATE \textbf{Return:}  $\mathcal{D}_{t}$,
    $\mathcal{I}_{t}$,
    $\mathcal{P}_{t}$
\end{algorithmic}
\end{algorithm}

\subsection{Image-level Prototypes Refinement}
To address the challenge of mitigating interference among tasks while preserving historical knowledge, our approach dynamically refines the image-level prototype for each task. 

The process begins by computing an raw prototype  $\mathbf{i}_\text{raw}$ as the average feature extracted from the current task’s data using a pretrained model $f_\theta$. Prototypes stored from previous tasks  $\mathcal{I}_{t-1}$ serve as reference anchors to guide the refinement of this raw representation.

Refinement is achieved by optimizing the image-level prototype refinement loss function $\mathcal{L}_{\text{IPR}}$ with two complementary objectives:


\begin{itemize}
    \item Similarity Preservation Loss $\mathcal{L}_{\text{sim}}$: Maximizes cosine similarity between the refined prototype $\mathbf{i}_\text{t}$ and the raw prototype $\mathbf{i}_\text{raw}$:

    \begin{equation}
        \mathcal{L}_{\text{sim}} = -\cos \left( \mathbf{i}_\text{t}, \mathbf{i}_\text{raw} \right).
    \end{equation}
    \item Inter-Task Distinction Loss $\mathcal{L}_{\text{dist}}$: Minimizes the maximum cosine similarity between $\mathbf{i}_\text{t}$ and the historical prototypes $\mathcal{I}_{t-1}$:
    \begin{equation}
        \mathcal{L}_{\text{dist}} = \max_{\mathbf{i} \in \mathcal{I}_{t-1}} \cos \left( \mathbf{i}_\text{t}, \mathbf{i} \right).
    \end{equation}
\end{itemize}
The total loss $\mathcal{L}_{\text{IPR}}=\mathcal{L}_{\text{sim}} + \mathcal{L}_{\text{dist}}$ is minimized using the L-BFGS optimizer~\cite{moritz2016linearly}. This produces the refined prototype $\mathbf{i}_\text{t}$, which updates the prototypes bank as $\mathcal{I}_{t}\gets\mathcal{I}_{t-1} \cup \{\mathbf{i}_\text{t}\}$. By balancing task-specific semantics and historical consistency, this dual-objective optimization mitigates interference across tasks.  

\subsection{Experience Replay-based Optimization}
To address the dual challenges of cross-task feature conflicts (where overlapping semantics across tasks degrade discriminability) and ambiguous pixel-level representations (which hinder precise anomaly localization), we propose a unified experience replay based optimization framework integrating inter-task separation and intra-task contrastive learning.

For inter-task conflicts, we explicitly decouple task-specific features through two complementary losses:
\begin{itemize}
    \item Image-Level Prototype Alignment Loss $\mathcal{L}_\text{align}$: 
     \begin{equation}
     \mathcal{L}_{\text{align}} = \left\| \mathbf{k}_\text{I} - \mathbf{i}_\text{t}\right\|_2
 \end{equation}
 minimizing the L2 distance between the current image-level feature $\mathbf{k}_\text{I}$ and the refined prototype $\mathbf{i}_\text{t}\in\mathcal{I}_t$.
    
    \item Inter-Task Contrastive Loss $ \mathcal{L}_{\text{ITC}}$: 
    \begin{equation}
        \mathcal{L}_{\text{ITC}} = \sum_{i\in N_p} \max_{\mathbf{p} \in \mathcal{P}_{t-1}} \cos \left(\mathbf{k}_\text{{P,i}}, \mathbf{p}\right),
    \end{equation}
    reducing similarity between current patch features $\mathbf{k}_\text{{P,i}}$ and historical pixel-level prototypes $\mathcal{P}_{t-1}$.
\end{itemize}

This dual objective explicitly separates task-specific features to prevent semantic overlap.

To refine intra-task discriminability, the Task-Specific Contrastive Loss $\mathcal{L}_{\text{TSC}}$ enforces class-aware constraints using SAM~\cite{kirillov2023segment} for patch-level class labels $C_{{P, i}}$: 
\begin{equation}
    \begin{aligned}
    &\mathcal{L}_{\text{pos}} = \sum_{i,j \in {N}_p} \cos \left (\mathbf{k}_\text{{P,i}}, \mathbf{k}_\text{{P, j}} \right), C_{{P, i}}=C_{P, j}, \\
    &\mathcal{L}_{\text{neg}} = \sum_{i,j \in {N}_p} \cos \left (\mathbf{k}_\text{{P, i}}, \mathbf{k}_\text{{P, j}} \right ), C_{{P, i}}\neq C_{{P, j}}, \\
    &\mathcal{L}_{\text{TSC}} = \mathcal{L}_{\text{pos}} - \mathcal{L}_{\text{neg}}.\\
    \end{aligned}
\end{equation}
By maximizing intra-class similarity ($\mathcal{L}_{\text{pos}}$) and minimizing inter-class similarity~($\mathcal{L}_{\text{neg}}$), the model learns compact representations for accurate anomaly localization.
\subsection{Incremental Selection of  Pixel-level Prototypes}
Existing prototype selection methods~\cite{Roth_2022_CVPR, liu2024unsupervised} prioritize intra-task representativeness but neglect inter-task discriminability, leading to cross-task semantic overlap. To overcome this, our approach jointly optimizes intra-task coverage and inter-task separation while maintaining storage efficiency.

After training task $t$, the pretrained model $f_\phi$ extracts pixel-level features from $\mathcal{T}_t$, forming a feature matrix $\mathbf{p}_\text{all} \in \mathbb{R}^{N_I \times N_p \times d}$, where $N_I$ is the number of training samples in $\mathcal{T}_t$, and $N_p$ is the number of patches per image.

To optimize storage and inference efficiency, we select $N_p$ prototypes~($\mathbf{p}_\text{t} \in \mathbf{p}_\text{all}$)~through a two-step strategy:

\begin{enumerate}
    \item Coreset Selection: Prototypes are selected by minimizing the maximum distance between any feature in $\mathbf{p}_\text{all}$ and the nearest prototype in either the current coreset $\mathcal{E}_C$ or historical prototypes $\mathcal{P}_{t-1}$:
    \begin{equation}
        \mathbf{p}_\text{t} = \arg\min_{\mathcal{E}_C \subseteq \mathbf{p}_\text{all}}  \max_{m \in \mathbf{p}_\text{all}} \min_{n \in \mathcal{E}_C\cup\mathcal{P}_{t-1}} \| m - n \|^2.
    \end{equation}
 
    \item Prototype Integration: The selected prototypes are merged into the global prototype bank:
   \begin{equation}
    \mathcal{P}_{t} \gets \mathcal{P}_{t-1}  \cup \{\mathbf{p}_\text{t}\}.
\end{equation}
\end{enumerate}
By enforcing both intra-task representativeness and inter-task separation, the prototype bank $\mathcal{P}$ retains task-specific semantics while preventing cross-task contamination, thereby enhancing incremental anomaly detection robustness.

\section{Experiments}
\begin{table*}[!t]
    \centering
    {\small
    \begin{tabular}{lccccccccc}
    \toprule
    \multirow{2}{*}{Method}& Extra info& 
    \multicolumn{2}{c}{Image-level ACC~$\uparrow$} & \multicolumn{2}{c}{Image-level FM~$\downarrow$}& 
    \multicolumn{2}{c}{Pixel-level ACC~$\uparrow$} & \multicolumn{2}{c}{Pixel-level FM~$\downarrow$}\\
       \cmidrule(lr){2-2} \cmidrule(lr){3-4} \cmidrule(lr){5-6}  \cmidrule(lr){7-8} \cmidrule(lr){9-10}
        ~&SR~$|$~TI~&AUROC & AUPR & AUROC & AUPR &AUROC & AUPR &  AUROC & AUPR\\
   \midrule
   \textit{Training-based ADs:}\\
        \hspace{1em}DRAEM~\textcolor{gray}{\tiny [ICCV21]} & $\times~|~\times$&57.7&78.7&39.4&20.0&61.6&11.7&34.3&53.6\\
        \hspace{1em}BGAD~\textcolor{gray}{\tiny [CVPR23]} & $\times~|~\times$ &51.8&73.9&42.1&22.6&50.5&6.6&46.0&48.0 \\
        \hspace{1em}DRAEM-Replay &$\checkmark~|~\times$&74.9&87.7&22.3&11.1&85.0&\underline{37.6}&10.8&27.7\\
        \hspace{1em}BGAD-Replay &$\checkmark~|~\times$&79.9&89.5&14.0&7.0&\underline{91.3}&33.0&5.1&21.7\\
        \hspace{1em}DRAEM-CLS* &$\checkmark~|~\times$&80.6&87.1&16.5&11.7&82.7&29.7&13.2&35.6\\
        \hspace{1em}BGAD-CLS* &$\checkmark~|~\times$&83.4&89.3&10.5&7.2&84.8&24.5&11.7&30.1\\
        \hline
   \textit{Training-free ADs:}\\
        \hspace{1em}WinCLIP \textcolor{gray}{\tiny [CVPR23]}&$\times~|~\checkmark$&90.4&95.6&/&/&82.3&18.2&/&/\\
        \hspace{1em}SAA++ \textcolor{gray}{\tiny [Arxiv23]}&$\times~|~\checkmark$&63.1&81.4&/&/&73.2&28.8&/&/\\
        \hspace{1em}CLIP-Surgery~\textcolor{gray}{\tiny [Arxiv23]} &$\times~|~\checkmark$&90.2&95.5&/&/&83.5&23.2&/&/\\

        \hspace{1em}FADE~\textcolor{gray}{\tiny [BMVC24]} & $\times~|~\checkmark$& 90.1&95.5&/&/&89.2&39.8&/&/\\
        \hspace{1em}SDP~\textcolor{gray}{\tiny [IJCAI24]} &$\times~|~\checkmark$&\underline{90.9}&\underline{95.8}&/&/&87.5&30.4&/&/\\
        \hspace{1em}PatchCore*~\textcolor{gray}{\tiny [CVPR22]} &$\times~|~\times$&90.4&94.9&\underline{0.7}&\underline{0.3}&89.7&22.1&\underline{3.8}&19.3\\
        \hline
       \textit{Incremental ADs:}\\

        \hspace{1em}CAD~\textcolor{gray}{\tiny [ACM MM23]}&$\times~|~\times$&87.1&93.3&1.9&1.7&/&/&/&/\\
        \hspace{1em}CAD~+~PANDA&$\times~|~\times$&47.8&74.1&36.0&18.5&/&/&/&/\\
       \hspace{1em}CAD~+~CutPaste & $\times~|~\times$&74.6&86.3&17.2&10.0&/&/&/&/ \\
        \hspace{1em}IUF~\textcolor{gray}{\tiny [ECCV24]} &$\times~|~\times$&71.3&83.9&3.9&1.1&79.0&16.0&5.7&\underline{6.7}\\
        \hspace{1em}UCAD*~\textcolor{gray}{\tiny [AAAI24]} &$\times~|~\times$&90.7&95.3&1.8&0.9&90.1&21.4&6.1&24.2\\
        \hspace{1em}\textbf{ONER} (Ours) &$\times~|~\times$&\textbf{93.4}&\textbf{97.3}&\textbf{0.6}&\textbf{0.1}&\textbf{94.5}&\textbf{49.7}&\textbf{0.9}&\textbf{1.3}\\
        \bottomrule
    \end{tabular}}
    \caption{{Comparison of Pixel-level and Image-level ACC and FM on the MVTec AD dataset ($N=15$) after training on the last subset. Note that * signifies staged approaches. Bold and underline highlight the best and second-best results, respectively.}}
    \label{tab:mvtec-sota}
\end{table*}
\begin{table*}[!t]
    \centering
    {\small
    \begin{tabular}{lccccccccc}
    \toprule
    \multirow{2}{*}{Method}& Extra info& 
    \multicolumn{2}{c}{Image-level ACC~$\uparrow$} & \multicolumn{2}{c}{Image-level FM~$\downarrow$}& 
    \multicolumn{2}{c}{Pixel-level ACC~$\uparrow$} & \multicolumn{2}{c}{Pixel-level FM~$\downarrow$}\\
       \cmidrule(lr){2-2} \cmidrule(lr){3-4} \cmidrule(lr){5-6}  \cmidrule(lr){7-8} \cmidrule(lr){9-10}
        ~&SR~$|$~TI&AUROC & AUPR & AUROC & AUPR &AUROC & AUPR &  AUROC & AUPR\\
   \midrule
   \textit{Training-based ADs:}\\
        \hspace{1em}DRAEM~\textcolor{gray}{\tiny [ICCV21]} & $\times~|~\times$&45.1&55.1&49.3&41.0&56.7&3.9&37.2&30.6 \\
        \hspace{1em}BGAD~\textcolor{gray}{\tiny [CVPR23]} & $\times~|~\times$ & 49.1&53.5&45.0&41.6&53.7&4.2&45.3&36.7\\
        \hspace{1em}DRAEM-Replay &$\checkmark~|~\times$&80.4&83.0&14.0&13,1&80.9&18.4&13.0&16.2\\
        \hspace{1em}BGAD-Replay &$\checkmark~|~\times$&79.6&81.3&14.5&13.8&89.3&20.8&9.5&20.1\\
        \hspace{1em}DRAEM-CLS* &$\checkmark~|~\times$&80.6&\underline{85.7}&13.8&10.4&82.2&\underline{24.9}&11.7&9.6\\
        \hspace{1em}BGAD-CLS* &$\checkmark~|~\times$&\underline{81.3}&82.8&12.8&12.3&87.7&18.9&11.1&22.0\\
        \hline
        \textit{Training-free ADs:}\\
        \hspace{1em}WinCLIP \textcolor{gray}{\tiny [CVPR23]}&$\times~|~\checkmark$&75.5&78.7&/&/&73.2&5.4&/&/\\
        \hspace{1em}SAA++ \textcolor{gray}{\tiny [Arxiv23]}&$\times~|~\checkmark$&71.1&77.3&/&/&74.0&22.4&/&/\\
        \hspace{1em}CLIP-Surgery~\textcolor{gray}{\tiny [Arxiv23]} &$\times~|~\checkmark$&76.8&80.2&/&/&85.0&10.3&/&/\\

        \hspace{1em}FADE~\textcolor{gray}{\tiny [BMVC24]} & $\times~|~\checkmark$&74.7&78.5&/&/&\textbf{91.0}&10.0&/&/ \\
        \hspace{1em}SDP~\textcolor{gray}{\tiny [IJCAI24]} &$\times~|~\checkmark$&78.6&81.5&/&/&88.1&12.2&/&/\\
        \hspace{1em}PatchCore*~\textcolor{gray}{\tiny [CVPR22]} &$\times~|~\times$&78.7&82.7&\textbf{1.0}&\textbf{0.8}&88.7&15.9&\textbf{2.2}&9.1\\
        \hline
        \textit{Incremental ADs:}\\
        \hspace{1em}CAD~\textcolor{gray}{\tiny [ACM MM23]}&$\times~|~\times$&64.7&70.7&13.8&11.3&/&/&/&/\\
        \hspace{1em}CAD~+~PANDA&$\times~|~\times$&55.6&65.9&19.3&12.8&/&/&/&/\\
        \hspace{1em}CAD~+~CutPaste & $\times~|~\times$& 65.9&71.0&13.5&11.2&/&/&/&/\\
        \hspace{1em}IUF~\textcolor{gray}{\tiny [ECCV24]} &$\times~|~\times$&65.9&85.2&\underline{2.6}&4.7&85.6&3.8&\underline{2.3}&\textbf{1.6}\\
        \hspace{1em}UCAD*~\textcolor{gray}{\tiny [AAAI24]} &$\times~|~\times$&75.7&79.2&10.8&8.9&86.9&10.1&5.9&19.8\\
        \hspace{1em}\textbf{ONER} (Ours) &$\times~|~\times$&\textbf{83.1}&\textbf{86.6}&3.9&\underline{2.7}&\underline{90.6}&\textbf{27.9}&2.4&\underline{1.8}\\
        \bottomrule
    \end{tabular}}
    \caption{{Comparison of Pixel-level and Image-level ACC and FM on the VisA dataset ($N=12$) after training on the last subset. Note that * signifies staged approaches. Bold and underline highlight the best and second-best results, respectively.}}
    \label{tab:visa-sota}
\end{table*}
\begin{figure*}[!ht]
\centering
\includegraphics[width=\textwidth]{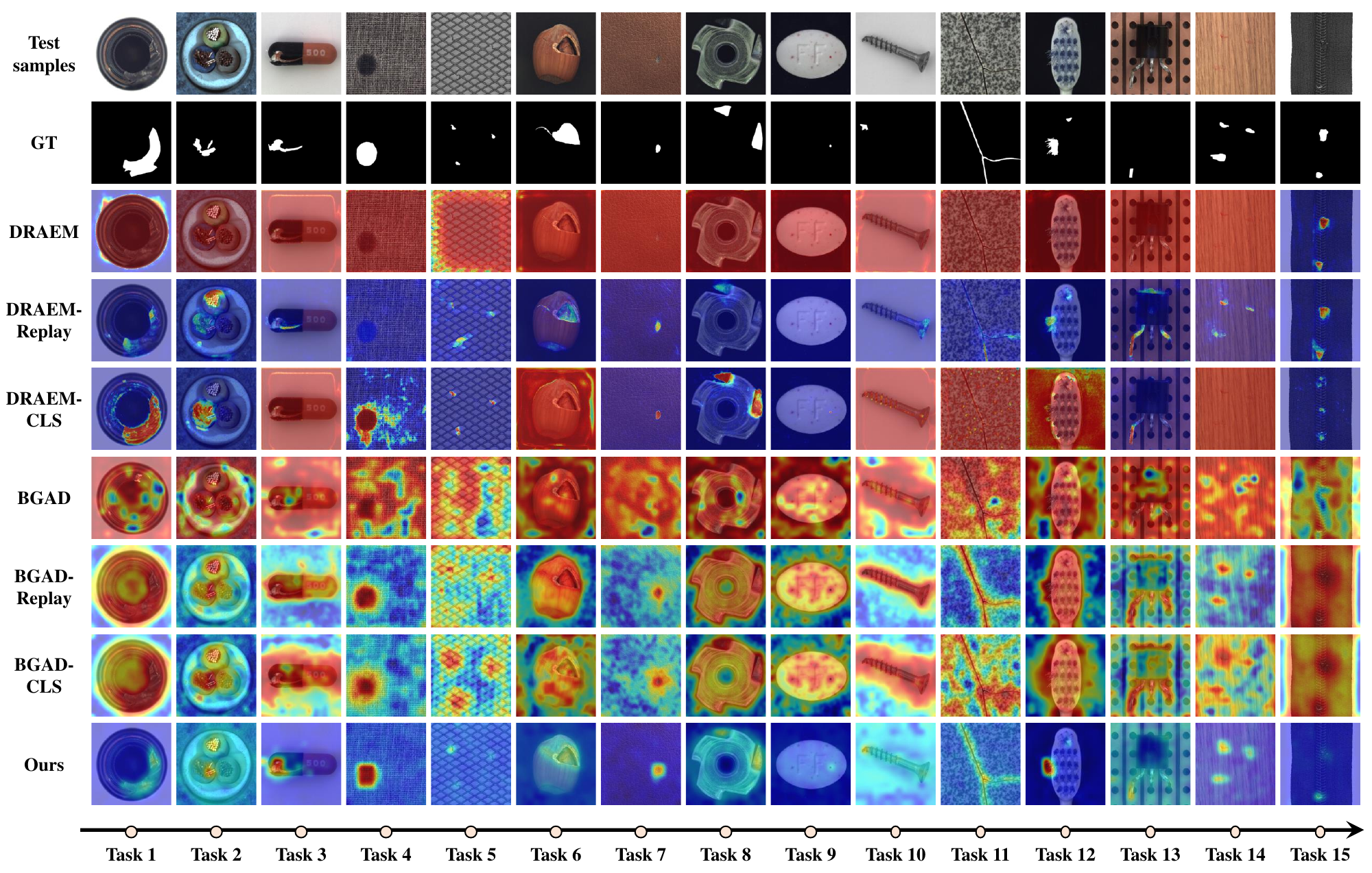} 
\caption{Qualitative evaluation on the MVTec AD dataset after training on the last subset. Experiments demonstrate our method's superior stability and knowledge retention over sample-replay (DRAEM-Replay, BGAD-Replay) and staged approaches (DRAEM-CLS, BGAD-CLS).}
\label{fig-replay}
\end{figure*}

\subsection{Experimental setup}
\textbf{Datasets}: Our experiments are conducted on two public datasets: MVTec AD \cite{Bergmann_2019_CVPR} and VisA \cite{zou2022spot}. MVTec AD includes 15 subsets and is one of the most widely used benchmarks for industrial anomaly detection in images. VisA contains 12 subsets and is one of the largest real-world industrial anomaly detection datasets, featuring pixel-level annotations.

\textbf{Comparison methods}: To comprehensively validate the effectiveness of our method, we conduct extensive comparisons against a variety of anomaly detection approaches, including training-based ADs, training-free ADs, and specialized incremental ADs.

For training-based ADs,  we compare our approach against both image reconstruction and feature embedding  based methods: DRAEM~\cite{zavrtanik2021draem} and BGAD~\cite{yao2023explicit}. To assess sample replay’s impact on catastrophic forgetting, we implement DRAEM-Replay and BGAD-Replay, where models are fine-tuned on the current task’s data and 5 replay samples per historical task. Inspired by staged approaches, we further design DRAEM-CLS and BGAD-CLS, integrating incremental classifiers trained via DER~\cite{buzzega2020dark} with task-specific models. For classifier training, we similarly retain 5 samples per historical task (see Appendix for implementation details).
For training-free ADs, we compare our approach with PatchCore~\cite{roth2022towards}, WinCLIP~\cite{jeong2023winclip}, FADE~\cite{li2024fade}, SAA++~\cite{cao2023segment}, CLIP-Surgery~\cite{li2023clip}, SDP~\cite{chen2024clip} and PatchCore, as a task-specific method, requires explicit memory bank selection during testing. To allow a fair comparison, we equip PatchCore with UCAD’s task identification module~\cite{liu2024unsupervised}. 
For incremental ADs, we compare our approach with UCAD~\cite{liu2024unsupervised}, IUF~\cite{tang2024incremental} and CAD~\cite{DBLP:conf/mm/LiZWXGLWZ22}. Notable, we additionally integrate PANDA~\cite{reiss2021panda} and CutPaste~\cite{li2021cutpaste} in CAD.

\textbf{Evaluation metrics}: We employ several metrics to comprehensively evaluate our method's performance. Specifically, we use AUROC, and AUPR as the evaluation metrics for both image-level and pixel-level anomaly detection. Notably, we report both Accuracy (ACC) and Forgetting Measure (FM) for each metric. The calculation of FM differs between staged approaches and end-to-end methods, as each mode has distinct mechanisms to handle catastrophic forgetting. Details on FM computation for both modes are provided in the Appendix.


\textbf{Implementation Details}: We utilize the vit-base-patch16-224 backbone pretrained on ImageNet for our method. During training, we employ a batch size of 8 and adapt Adam optimizer with a learning rate of 0.0005 and momentum of 0.9. The training process spanned 5 epochs.
\subsection{Incremental anomaly detection benchmark}
We conduct comprehensive evaluations of aforementioned methods on the MVTec AD and VisA datasets. Among them, DRAEM-REPLAY, BGAD-REPLAY, DRAEM-CLS, and BGAD-CLS employ sample replay (SR) to mitigate catastrophic forgetting. Additionally, WinCLIP, SAA++, CLIP-Surgergy, FADE and SDP leverage textual information (TI). These evaluations provide a thorough comparison across diverse methodologies, highlighting the strengths and limitations of each approach.

\subsubsection{Quantitative Analysis}

\begin{table}[t]
\centering
\small
\begin{tabular}{lcc}
\toprule
Method & Learnable Parameters & Epochs \\
\midrule
DRAEM~\textcolor{gray}{\tiny [ICCV21]} & 97.42M & 300 \\
BGAD~\textcolor{gray}{\tiny [CVPR23]} & 3.39M & 100 \\
DRAEM-Replay & 97.42M & 300\\
BGAD-Replay & 3.39M & 100\\
DRAEM-CLS* & (97.42+8.40)M & 300+10\\
BGAD-CLS* & (3.39+8.40)M & 100+10\\
\hline

CAD~\textcolor{gray}{\tiny [ACM MM23]} & 85.21M & 50 \\
IUF~\textcolor{gray}{\tiny [ECCV24]} & 9.11M & 200 \\
UCAD*~\textcolor{gray}{\tiny [AAAI24]} & 0.021M & 25 \\

\hline
\textbf{ONER} (Ours) & \textbf{0.019M} & \textbf{5} \\
\bottomrule
\end{tabular}
\caption{Statistics of learnable parameters and training epochs per task for different methods on the MVTec AD dataset.}
\label{tab:parameter_epochs_statistics}

\end{table}
\begin{table}[b]
{\footnotesize
\begin{tabular}{lccccc}
\toprule
\textbf{ver.} & \textbf{Method} & \multicolumn{2}{c}{\textbf{Image-level}} & \multicolumn{2}{c}{\textbf{Pixel-level}} \\
\cmidrule(lr){3-4} \cmidrule(lr){5-6}
 &  & \textbf{AUROC} & \textbf{AUPR} & \textbf{AUROC} & \textbf{AUPR} \\
\midrule
1.0 & Baseline & 77.87 & 82.44 & 87.63 & 23.83 \\
\midrule
2.0 & 1.0 + $\mathcal{D}$  + $\mathcal{L}_\text{TSC}$ & 79.11 & 81.94 & 88.44 & 24.89 \\
\midrule
3.0 & 2.0 + $\mathcal{L}_\text{align}$  & 80.03 & 83.46 & 87.80 & 25.53 \\
3.1 & 3.0 + IPR  & 81.82 & 84.77 & 90.37 & 27.39 \\
\midrule
4.0 & 3.1 + $\mathcal{L}_\text{ITC}$  & 82.05 & 85.10 & 90.56 & 27.39 \\
4.1 & 4.0 + ISPP & \textbf{83.14} & \textbf{86.61} & \textbf{90.58} & \textbf{27.87} \\
\bottomrule
\end{tabular}}
\caption{Ablation study of our method on VisA Dataset. The table shows the impact of incrementally adding Experience replay by Decomposed Prompts (ver.~2.0), Image-level Prototypes (ver.~3.0-3.1), and Pixel-level Prototypes (ver.~4.0-4.1) on both image-level and pixel-level AUROC and AUPR.}
\label{tab:ablation_study}

\end{table}
As shown in Tab.~\ref{tab:mvtec-sota}--\ref{tab:visa-sota}, our proposed method achieves state-of-the-art performance on both the MVTec AD and VisA datasets. On the MVTec AD, ONER surpasses all competing methods, demonstrating substantial gains in pixel-level anomaly detection while maintaining minimal forgetting.

We observe that sample replay significantly alleviates catastrophic forgetting, as evidenced by the improved performance of DRAEM-REPLAY and BGAD-REPLAY over their vanilla counterparts. However, this strategy requires the storage and replaying of historical data, rendering it impractical for data-restricted real-world scenarios. Similarly, while staged approaches like DRAEM-CLS and BGAD-CLS achieve competitive results, they are prone to error accumulation across tasks, resulting in unstable performance over time.

On the challenging VisA dataset, PatchCore* achieves a lower image-level forgetting measure (FM) but fails to integrate task-aware knowledge effectively, yielding a pixel-level AUPR of only 15.9\%. Its staged design further exacerbates error accumulation, causing unstable detection results. Incontrast, ONER achieves state-of-the-art results (83.1\% image-level AUROC, 86.6\% image-level AUPR and 27.9\% pixel-level AUPR). Remarkably, this is accomplished without sample replay, while maintaining minimal training costs (only 0.019M learnable parameters and 5 training epochs per task, as shown in Tab.~\ref{tab:parameter_epochs_statistics}).

In summary, ONER strikes an optimal balance between performance, efficiency, and stability, making it a practical solution for real-world industrial applications with dynamic and resource-constrained environments.

\subsubsection{Qualitative Analysis}
As shown in Fig.~\ref{fig-replay}, our experiments demonstrate that sample replay effectively mitigates catastrophic forgetting, with DRAEM-REPLAY and BGAD-REPLAY outperforming their vanilla counterparts. In contrast, our method achieves better results with replaying only representative features, which is memory-efficient and privacy preserving while effectively resisting forgetting. Staged aproaches like DRAEM-CLS and BGAD-CLS, though competitive, exhibit critical instability: inaccurate task identification (e.g., in Task 3 and Task 6) triggers substantial performance drops due to error accumulation, highlighting its unreliability. In contrast, our method directly processes input images, resulting in more stable and reliable detection outcomes. Additional examples are provided in the Appendix.

\subsubsection{Ablation Study}
An improvement in the model's performance is noted with the incorporation of the decomposed prompt experience (ver.~2.0).
However, the model may suffer from distribution shift since the differences of inter-task feature are unconstrained, leading to a decline of image-level AUPR. 
The continuous improvement in model's performance can be observed with the implementation of $\mathcal{L}_\text{align}$ (ver.~3.0), IPR (ver.~3.1), $\mathcal{L}_\text{ITC}$ (ver.~4.0) and ISPP (ver.~4.1). The ablation study demonstrates the effectiveness of proposed key components in anomaly detection.

\section{Conclusion}
To conclude, our proposed ONER method effectively tackles the challenges of distribution shift and catastrophic forgetting in incremental anomaly detection. By performing experience replay with decomposed prompts and semantic prototypes, ONER preserves high accuracy across tasks while enhancing feature discriminability. Experimental results validate that ONER not only achieves state-of-the-art performance but also demonstrates remarkable adaptability to new categories with minimal training overheads, making it a stable and efficient solution for real-world applications.

{
    \small
    \bibliographystyle{ieeenat_fullname}
    \bibliography{main_new}

\begin{thebibliography}{55}
\providecommand{\natexlab}[1]{#1}
\providecommand{\url}[1]{\texttt{#1}}
\expandafter\ifx\csname urlstyle\endcsname\relax
  \providecommand{\doi}[1]{doi: #1}\else
  \providecommand{\doi}{doi: \begingroup \urlstyle{rm}\Url}\fi

\bibitem[Audibert et~al.(2020)Audibert, Michiardi, Guyard, Marti, and Zuluaga]{DBLP:conf/kdd/AudibertMGMZ20}
Julien Audibert, Pietro Michiardi, Fr{\'{e}}d{\'{e}}ric Guyard, S{\'{e}}bastien Marti, and Maria~A. Zuluaga.
\newblock {USAD:} unsupervised anomaly detection on multivariate time series.
\newblock In \emph{{KDD} '20: The 26th {ACM} {SIGKDD} Conference on Knowledge Discovery and Data Mining, Virtual Event, CA, USA, August 23-27, 2020}, pages 3395--3404. {ACM}, 2020.

\bibitem[Bergmann et~al.(2019{\natexlab{a}})Bergmann, Fauser, Sattlegger, and Steger]{Bergmann_2019_CVPR}
Paul Bergmann, Michael Fauser, David Sattlegger, and Carsten Steger.
\newblock Mvtec ad -- a comprehensive real-world dataset for unsupervised anomaly detection.
\newblock In \emph{CVPR}, 2019{\natexlab{a}}.

\bibitem[Bergmann et~al.(2019{\natexlab{b}})Bergmann, Löwe, Fauser, Sattlegger, and Steger]{Bergmann_Löwe_Fauser_Sattlegger_Steger_2019}
Paul Bergmann, Sindy Löwe, Michael Fauser, David Sattlegger, and Carsten Steger.
\newblock Improving unsupervised defect segmentation by applying structural similarity to autoencoders.
\newblock In \emph{Proceedings of the 14th International Joint Conference on Computer Vision, Imaging and Computer Graphics Theory and Applications}, 2019{\natexlab{b}}.

\bibitem[Buzzega et~al.(2020)Buzzega, Boschini, Porrello, Abati, and Calderara]{buzzega2020dark}
Pietro Buzzega, Matteo Boschini, Angelo Porrello, Davide Abati, and Simone Calderara.
\newblock Dark experience for general continual learning: a strong, simple baseline.
\newblock \emph{NeurIPS}, 33:\penalty0 15920--15930, 2020.

\bibitem[Cao et~al.(2023)Cao, Xu, Sun, Cheng, Du, Gao, and Shen]{cao2023segment}
Yunkang Cao, Xiaohao Xu, Chen Sun, Yuqi Cheng, Zongwei Du, Liang Gao, and Weiming Shen.
\newblock Segment any anomaly without training via hybrid prompt regularization, 2023.
\newblock arXiv preprint arXiv:2305.10724.

\bibitem[Chen et~al.(2024)Chen, Zhang, Tian, He, Zhang, Wang, Wang, and Liu]{chen2024clip}
Xuhai Chen, Jiangning Zhang, Guanzhong Tian, Haoyang He, Wuhao Zhang, Yabiao Wang, Chengjie Wang, and Yong Liu.
\newblock Clip-ad: A language-guided staged dual-path model for zero-shot anomaly detection.
\newblock In \emph{IJCAI}, pages 17--33, 2024.

\bibitem[Cohen and Hoshen(2020)]{cohen2020sub}
Niv Cohen and Yedid Hoshen.
\newblock Sub-image anomaly detection with deep pyramid correspondences, 2020.
\newblock arXiv preprint arXiv:2005.02357.

\bibitem[Dosovitskiy(2020)]{dosovitskiy2020image}
Alexey Dosovitskiy.
\newblock An image is worth 16x16 words: Transformers for image recognition at scale, 2020.
\newblock arXiv preprint arXiv:2010.11929.

\bibitem[Gao et~al.(2023)Gao, Luo, Shen, and Zhang]{DBLP:journals/corr/abs-2305-15652}
Han Gao, Huiyuan Luo, Fei Shen, and Zhengtao Zhang.
\newblock Towards total online unsupervised anomaly detection and localization in industrial vision.
\newblock \emph{CoRR}, abs/2305.15652, 2023.

\bibitem[He et~al.(2024)He, Zhang, Chen, Chen, Li, Chen, Wang, Wang, and Xie]{DBLP:conf/aaai/HeZCCLCWW024}
Haoyang He, Jiangning Zhang, Hongxu Chen, Xuhai Chen, Zhishan Li, Xu Chen, Yabiao Wang, Chengjie Wang, and Lei Xie.
\newblock A diffusion-based framework for multi-class anomaly detection.
\newblock In \emph{AAAI}, pages 8472--8480, 2024.

\bibitem[Jeong et~al.(2023)Jeong, Zou, Kim, Zhang, Ravichandran, and Dabeer]{jeong2023winclip}
Jongheon Jeong, Yang Zou, Taewan Kim, Dongqing Zhang, Avinash Ravichandran, and Onkar Dabeer.
\newblock Winclip: Zero-/few-shot anomaly classification and segmentation.
\newblock In \emph{CVPR}, pages 19606--19616, 2023.

\bibitem[Kim et~al.(2023)Kim, Park, Cho, and Lee]{kim2023fapm}
Donghyeong Kim, Chaewon Park, Suhwan Cho, and Sangyoun Lee.
\newblock Fapm: Fast adaptive patch memory for real-time industrial anomaly detection.
\newblock In \emph{ICASSP}, pages 1--5. IEEE, 2023.

\bibitem[Kim et~al.(2021)Kim, Kim, Yi, and Lee]{kim2021semi}
Jin-Hwa Kim, Do-Hyeong Kim, Saehoon Yi, and Taehoon Lee.
\newblock Semi-orthogonal embedding for efficient unsupervised anomaly segmentation, 2021.
\newblock arXiv preprint arXiv:2105.14737.

\bibitem[Kirillov et~al.(2023)Kirillov, Mintun, Ravi, Mao, Rolland, Gustafson, Xiao, Whitehead, Berg, Lo, et~al.]{kirillov2023segment}
Alexander Kirillov, Eric Mintun, Nikhila Ravi, Hanzi Mao, Chloe Rolland, Laura Gustafson, Tete Xiao, Spencer Whitehead, Alexander~C Berg, Wan-Yen Lo, et~al.
\newblock Segment anything.
\newblock In \emph{ICCV}, pages 4015--4026, 2023.

\bibitem[Li et~al.(2021{\natexlab{a}})Li, Sohn, Yoon, and Pfister]{DBLP:conf/cvpr/LiSYP21}
Chun{-}Liang Li, Kihyuk Sohn, Jinsung Yoon, and Tomas Pfister.
\newblock Cutpaste: Self-supervised learning for anomaly detection and localization.
\newblock In \emph{CVPR}, pages 9664--9674, 2021{\natexlab{a}}.

\bibitem[Li et~al.(2021{\natexlab{b}})Li, Sohn, Yoon, and Pfister]{li2021cutpaste}
Chun-Liang Li, Kihyuk Sohn, Jinsung Yoon, and Tomas Pfister.
\newblock Cutpaste: Self-supervised learning for anomaly detection and localization.
\newblock In \emph{CVPR}, pages 9664--9674, 2021{\natexlab{b}}.

\bibitem[Li et~al.(2022{\natexlab{a}})Li, Zhan, Wang, Xia, Gao, Liu, Wang, and Zheng]{DBLP:conf/mm/LiZWXGLWZ22}
Wujin Li, Jiawei Zhan, Jinbao Wang, Bizhong Xia, Bin{-}Bin Gao, Jun Liu, Chengjie Wang, and Feng Zheng.
\newblock Towards continual adaptation in industrial anomaly detection.
\newblock In \emph{ACM MM}, pages 2871--2880. {ACM}, 2022{\natexlab{a}}.

\bibitem[Li et~al.(2022{\natexlab{b}})Li, Zhan, Wang, Xia, Gao, Liu, Wang, and Zheng]{li2022towards}
Wujin Li, Jiawei Zhan, Jinbao Wang, Bizhong Xia, Bin-Bin Gao, Jun Liu, Chengjie Wang, and Feng Zheng.
\newblock Towards continual adaptation in industrial anomaly detection.
\newblock In \emph{ACM MM}, pages 2871--2880, 2022{\natexlab{b}}.

\bibitem[Li et~al.(2023)Li, Wang, Duan, and Li]{li2023clip}
Yi Li, Hualiang Wang, Yiqun Duan, and Xiaomeng Li.
\newblock Clip surgery for better explainability with enhancement in open-vocabulary tasks, 2023.
\newblock arXiv preprint arXiv:2304.10293.

\bibitem[Li et~al.(2024{\natexlab{a}})Li, Ivanova, and Bruveris]{li2024fade}
Yuanwei Li, Elizaveta Ivanova, and Martins Bruveris.
\newblock Fade: Few-shot/zero-shot anomaly detection engine using large vision-language model.
\newblock In \emph{BMVC}, 2024{\natexlab{a}}.

\bibitem[Li et~al.(2024{\natexlab{b}})Li, Ivanova, and London]{Li_Ivanova_London}
Yuanwei Li, Elizaveta Ivanova, and Onfido London.
\newblock Fade: Few-shot/zero-shot anomaly detection engine using large vision-language model.
\newblock \emph{BMVC}, 2024{\natexlab{b}}.

\bibitem[Liang et~al.(2023)Liang, Zhang, Zhao, Wu, Liu, and Pan]{liang2023omni}
Yufei Liang, Jiangning Zhang, Shiwei Zhao, Runze Wu, Yong Liu, and Shuwen Pan.
\newblock Omni-frequency channel-selection representations for unsupervised anomaly detection.
\newblock \emph{IEEE TIP}, 32:\penalty0 4327--4340, 2023.

\bibitem[Liu et~al.(2024{\natexlab{a}})Liu, Wu, Nie, Chen, Gao, Liu, Wang, Wang, and Zheng]{DBLP:conf/aaai/0004WNCGLWWZ24}
Jiaqi Liu, Kai Wu, Qiang Nie, Ying Chen, Bin{-}Bin Gao, Yong Liu, Jinbao Wang, Chengjie Wang, and Feng Zheng.
\newblock Unsupervised continual anomaly detection with contrastively-learned prompt.
\newblock In \emph{AAAI}, pages 3639--3647. {AAAI} Press, 2024{\natexlab{a}}.

\bibitem[Liu et~al.(2024{\natexlab{b}})Liu, Wu, Nie, Chen, Gao, Liu, Wang, Wang, and Zheng]{liu2024unsupervised}
Jiaqi Liu, Kai Wu, Qiang Nie, Ying Chen, Bin-Bin Gao, Yong Liu, Jinbao Wang, Chengjie Wang, and Feng Zheng.
\newblock Unsupervised continual anomaly detection with contrastively-learned prompt.
\newblock In \emph{AAAI}, pages 3639--3647, 2024{\natexlab{b}}.

\bibitem[Liu et~al.(2024{\natexlab{c}})Liu, Xie, Wang, Li, Wang, Zheng, and Jin]{LiuXWLWZJ24}
Jiaqi Liu, Guoyang Xie, Jin{-}Bao Wang, Shangnian Li, Chengjie Wang, Feng Zheng, and Yaochu Jin.
\newblock Deep industrial image anomaly detection: {A} survey.
\newblock \emph{Mach. Intell. Res.}, 21\penalty0 (1):\penalty0 104--135, 2024{\natexlab{c}}.

\bibitem[Moritz et~al.(2016)Moritz, Nishihara, and Jordan]{moritz2016linearly}
Philipp Moritz, Robert Nishihara, and Michael Jordan.
\newblock A linearly-convergent stochastic l-bfgs algorithm.
\newblock In \emph{Artificial Intelligence and Statistics}, pages 249--258. PMLR, 2016.

\bibitem[Qu et~al.(2024)Qu, Tao, Prasad, Shen, Zhang, Gong, and Ding]{qu2024vcp}
Zhen Qu, Xian Tao, Mukesh Prasad, Fei Shen, Zhengtao Zhang, Xinyi Gong, and Guiguang Ding.
\newblock Vcp-clip: A visual context prompting model for zero-shot anomaly segmentation.
\newblock In \emph{ECCV}, pages 301--317. Springer, 2024.

\bibitem[Radford et~al.(2021)Radford, Kim, Hallacy, Ramesh, Goh, Agarwal, Sastry, Askell, Mishkin, Clark, et~al.]{radford2021learning}
Alec Radford, Jong~Wook Kim, Chris Hallacy, Aditya Ramesh, Gabriel Goh, Sandhini Agarwal, Girish Sastry, Amanda Askell, Pamela Mishkin, Jack Clark, et~al.
\newblock Learning transferable visual models from natural language supervision.
\newblock In \emph{International conference on machine learning}, pages 8748--8763. PmLR, 2021.

\bibitem[Reiss et~al.(2021{\natexlab{a}})Reiss, Cohen, Bergman, and Hoshen]{DBLP:conf/cvpr/ReissCBH21}
Tal Reiss, Niv Cohen, Liron Bergman, and Yedid Hoshen.
\newblock {PANDA:} adapting pretrained features for anomaly detection and segmentation.
\newblock In \emph{CVPR}, pages 2806--2814, 2021{\natexlab{a}}.

\bibitem[Reiss et~al.(2021{\natexlab{b}})Reiss, Cohen, Bergman, and Hoshen]{reiss2021panda}
Tal Reiss, Niv Cohen, Liron Bergman, and Yedid Hoshen.
\newblock Panda: Adapting pretrained features for anomaly detection and segmentation.
\newblock In \emph{CVPR}, pages 2806--2814, 2021{\natexlab{b}}.

\bibitem[Rolnick et~al.(2019)Rolnick, Ahuja, Schwarz, Lillicrap, and Wayne]{rolnick2019experience}
David Rolnick, Arun Ahuja, Jonathan Schwarz, Timothy Lillicrap, and Gregory Wayne.
\newblock Experience replay for continual learning.
\newblock \emph{NeurIPS}, 32, 2019.

\bibitem[Roth et~al.(2022{\natexlab{a}})Roth, Pemula, Zepeda, Sch\"olkopf, Brox, and Gehler]{Roth_2022_CVPR}
Karsten Roth, Latha Pemula, Joaquin Zepeda, Bernhard Sch\"olkopf, Thomas Brox, and Peter Gehler.
\newblock Towards total recall in industrial anomaly detection.
\newblock In \emph{CVPR}, pages 14318--14328, 2022{\natexlab{a}}.

\bibitem[Roth et~al.(2022{\natexlab{b}})Roth, Pemula, Zepeda, Sch{\"o}lkopf, Brox, and Gehler]{roth2022towards}
Karsten Roth, Latha Pemula, Joaquin Zepeda, Bernhard Sch{\"o}lkopf, Thomas Brox, and Peter Gehler.
\newblock Towards total recall in industrial anomaly detection.
\newblock In \emph{CVPR}, pages 14318--14328, 2022{\natexlab{b}}.

\bibitem[Rudolph et~al.(2021)Rudolph, Wandt, and Rosenhahn]{rudolph2021same}
Marco Rudolph, Bastian Wandt, and Bodo Rosenhahn.
\newblock Same same but differnet: Semi-supervised defect detection with normalizing flows.
\newblock In \emph{Proceedings of the IEEE/CVF winter conference on applications of computer vision}, pages 1907--1916, 2021.

\bibitem[Rudolph et~al.(2022)Rudolph, Wehrbein, Rosenhahn, and Wandt]{DBLP:conf/wacv/RudolphWRW22}
Marco Rudolph, Tom Wehrbein, Bodo Rosenhahn, and Bastian Wandt.
\newblock Fully convolutional cross-scale-flows for image-based defect detection.
\newblock In \emph{WACV}, pages 1088--1097, 2022.

\bibitem[Salehi et~al.(2021)Salehi, Sadjadi, Baselizadeh, Rohban, and Rabiee]{DBLP:conf/cvpr/SalehiSBRR21}
Mohammadreza Salehi, Niousha Sadjadi, Soroosh Baselizadeh, Mohammad~H. Rohban, and Hamid~R. Rabiee.
\newblock Multiresolution knowledge distillation for anomaly detection.
\newblock In \emph{CVPR}, pages 14902--14912, 2021.

\bibitem[Smith et~al.(2023)Smith, Karlinsky, Gutta, Cascante-Bonilla, Kim, Arbelle, Panda, Feris, and Kira]{smith2023coda}
James~Seale Smith, Leonid Karlinsky, Vyshnavi Gutta, Paola Cascante-Bonilla, Donghyun Kim, Assaf Arbelle, Rameswar Panda, Rogerio Feris, and Zsolt Kira.
\newblock Coda-prompt: Continual decomposed attention-based prompting for rehearsal-free continual learning.
\newblock In \emph{CVPR}, pages 11909--11919, 2023.

\bibitem[Song et~al.(2021)Song, Kong, Park, Kim, and Kang]{DBLP:journals/corr/abs-2110-03396}
Jou~Won Song, Kyeongbo Kong, Ye~In Park, Seong~Gyun Kim, and Suk{-}Ju Kang.
\newblock Anoseg: Anomaly segmentation network using self-supervised learning.
\newblock \emph{CoRR}, abs/2110.03396, 2021.

\bibitem[Tang et~al.(2024)Tang, Lu, Xu, Wu, Hu, Zhang, Cheng, Ge, Chen, and Tsung]{tang2024incremental}
Jiaqi Tang, Hao Lu, Xiaogang Xu, Ruizheng Wu, Sixing Hu, Tong Zhang, Tsz~Wa Cheng, Ming Ge, Ying-Cong Chen, and Fugee Tsung.
\newblock An incremental unified framework for small defect inspection.
\newblock In \emph{ECCV}, pages 307--324. Springer, 2024.

\bibitem[Towill et~al.(1997)Towill, Evans, and Cheema]{towill1997analysis}
Denis~R Towill, Gary~N Evans, and P Cheema.
\newblock Analysis and design of an adaptive minimum reasonable inventory control system.
\newblock \emph{Production Planning \& Control}, 8\penalty0 (6):\penalty0 545--557, 1997.

\bibitem[Wang et~al.(2021)Wang, Han, Ding, and Huang]{DBLP:conf/bmvc/WangHD021}
Guodong Wang, Shumin Han, Errui Ding, and Di Huang.
\newblock Student-teacher feature pyramid matching for anomaly detection.
\newblock In \emph{BMVC}, page 306, 2021.

\bibitem[Wang et~al.(2022{\natexlab{a}})Wang, Zhang, Ebrahimi, Sun, Zhang, Lee, Ren, Su, Perot, Dy, et~al.]{wang2022dualprompt}
Zifeng Wang, Zizhao Zhang, Sayna Ebrahimi, Ruoxi Sun, Han Zhang, Chen-Yu Lee, Xiaoqi Ren, Guolong Su, Vincent Perot, Jennifer Dy, et~al.
\newblock Dualprompt: Complementary prompting for rehearsal-free continual learning.
\newblock In \emph{ECCV}, pages 631--648. Springer, 2022{\natexlab{a}}.

\bibitem[Wang et~al.(2022{\natexlab{b}})Wang, Zhang, Lee, Zhang, Sun, Ren, Su, Perot, Dy, and Pfister]{Wang_Zhang_Lee_Zhang_Sun_Ren_Su_Perot_Dy_Pfister_2022}
Zifeng Wang, Zizhao Zhang, Chen-Yu Lee, Han Zhang, Ruoxi Sun, Xiaoqi Ren, Guolong Su, Vincent Perot, Jennifer Dy, and Tomas Pfister.
\newblock Learning to prompt for continual learning.
\newblock In \emph{CVPR}, 2022{\natexlab{b}}.

\bibitem[Wyatt et~al.(2022)Wyatt, Leach, Schmon, and Willcocks]{DBLP:conf/cvpr/WyattLSW22}
Julian Wyatt, Adam Leach, Sebastian~M. Schmon, and Chris~G. Willcocks.
\newblock Anoddpm: Anomaly detection with denoising diffusion probabilistic models using simplex noise.
\newblock In \emph{CVPR}, pages 649--655, 2022.

\bibitem[Xu et~al.(2024)Xu, Zhang, Li, Peng, and Zhou]{xu2024mitigate}
Kunlun Xu, Haozhuo Zhang, Yu Li, Yuxin Peng, and Jiahuan Zhou.
\newblock Mitigate catastrophic remembering via continual knowledge purification for noisy lifelong person re-identification.
\newblock In \emph{ACM MM}, pages 5790--5799, 2024.

\bibitem[Yan et~al.(2022)Yan, Zhang, Huang, Liu, Hu, Li, Liu, Jiang, Guo, and Zheng]{DBLP:journals/corr/abs-2206-01992}
Ruiqing Yan, Fan Zhang, Mengyuan Huang, Wu Liu, Dongyu Hu, Jinfeng Li, Qiang Liu, Jingrong Jiang, Qianjin Guo, and Linghan Zheng.
\newblock Cainnflow: Convolutional block attention modules and invertible neural networks flow for anomaly detection and localization tasks.
\newblock \emph{CoRR}, abs/2206.01992, 2022.

\bibitem[Yan et~al.(2021)Yan, Zhang, Xu, Hu, and Heng]{DBLP:conf/aaai/YanZXHH21}
Xudong Yan, Huaidong Zhang, Xuemiao Xu, Xiaowei Hu, and Pheng{-}Ann Heng.
\newblock Learning semantic context from normal samples for unsupervised anomaly detection.
\newblock In \emph{AAAI}, pages 3110--3118, 2021.

\bibitem[Yang et~al.(2023)Yang, Wu, and Feng]{DBLP:journals/eaai/YangWF23}
Minghui Yang, Peng Wu, and Hui Feng.
\newblock Memseg: {A} semi-supervised method for image surface defect detection using differences and commonalities.
\newblock \emph{Eng. Appl. Artif. Intell.}, 119:\penalty0 105835, 2023.

\bibitem[Yao et~al.(2024)Yao, Liu, Wang, Yin, Yan, Hong, and Zuo]{Yao_Liu_Wang_Yin_Yan_Hong_Zuo_2024}
Hang Yao, Ming Liu, Haolin Wang, Zhicun Yin, Zifei Yan, Xiaopeng Hong, and Wangmeng Zuo.
\newblock Glad: Towards better reconstruction with global and local adaptive diffusion models for unsupervised anomaly detection.
\newblock \emph{ECCV}, 2024.

\bibitem[Yao et~al.(2023)Yao, Li, Zhang, Sun, and Zhang]{yao2023explicit}
Xincheng Yao, Ruoqi Li, Jing Zhang, Jun Sun, and Chongyang Zhang.
\newblock Explicit boundary guided semi-push-pull contrastive learning for supervised anomaly detection.
\newblock In \emph{CVPR}, pages 24490--24499, 2023.

\bibitem[Yi and Yoon(2020)]{DBLP:conf/accv/YiY20}
Jihun Yi and Sungroh Yoon.
\newblock Patch {SVDD:} patch-level {SVDD} for anomaly detection and segmentation.
\newblock In \emph{ACCV}, pages 375--390. Springer, 2020.

\bibitem[Zavrtanik et~al.(2021{\natexlab{a}})Zavrtanik, Kristan, and Skocaj]{DBLP:conf/iccv/ZavrtanikKS21}
Vitjan Zavrtanik, Matej Kristan, and Danijel Skocaj.
\newblock Dr{\ae}m - {A} discriminatively trained reconstruction embedding for surface anomaly detection.
\newblock In \emph{ICCV}, pages 8310--8319. IEEE, 2021{\natexlab{a}}.

\bibitem[Zavrtanik et~al.(2021{\natexlab{b}})Zavrtanik, Kristan, and Sko{\v{c}}aj]{zavrtanik2021draem}
Vitjan Zavrtanik, Matej Kristan, and Danijel Sko{\v{c}}aj.
\newblock Draem-a discriminatively trained reconstruction embedding for surface anomaly detection.
\newblock In \emph{ICCV}, pages 8330--8339, 2021{\natexlab{b}}.

\bibitem[Zhou et~al.(2024)Zhou, Pang, Tian, He, and Chen]{Zhou_Pang_Tian_He_Chen_2024}
Qihang Zhou, Guansong Pang, Yu Tian, Shibo He, and Jiming Chen.
\newblock Anomalyclip: Object-agnostic prompt learning for zero-shot anomaly detection.
\newblock \emph{ICLR}, 2024.

\bibitem[Zou et~al.(2022)Zou, Jeong, Pemula, Zhang, and Dabeer]{zou2022spot}
Yang Zou, Jongheon Jeong, Latha Pemula, Dongqing Zhang, and Onkar Dabeer.
\newblock Spot-the-difference self-supervised pre-training for anomaly detection and segmentation.
\newblock In \emph{ECCV}, pages 392--408. Springer, 2022.

\end{thebibliography}
}

\end{document}